# A 96pJ/Frame/Pixel and 61pJ/Event Anti-UAV System with Hybrid Object Tracking Modes

Yuncheng Lu[1], Yucen Shi[1], Aobo Li[1], Zehao Li[1], Junying Li[1], Bo Wang[2], Tony Tae-Hyoung Kim[1]

[1]Nanyang Technological University, Singapore

[2]Singapore University of Technology and Design, Singapore

As UAVs proliferate in both civil and military applications, anti-UAV systems have become critical for safeguarding privacy and security. Due to the UAV's small size and varied speed range, vision-based detection is preferred over radar as it offers longer detection ranges and better small-target sensitivity. Event cameras, featuring sub-millisecond latency, excel in high-speed object detection but lack texture details essential for accurate classification. Recent work [1] addresses this by integrating RGB and event sensors (Fig. 1, top), combining a CNN-based approach for RGB frames with a Spiking Neural Network (SNN) for refining detections from event streams. However, two major limitations persist: existing systems typically evaluate slow drones within 50m, experiencing degraded performance with increased target speed and reduced size; additionally, continuously active neural networks result in excessive power consumption.

To tackle these challenges, we propose an energy-efficient anti-UAV system optimized for rapid detection of small, fast-moving objects. Fig. 1 (bottom) shows our hierarchical processing pipeline, which significantly reduces neural network (NN) computations through optimized preprocessing. Initially, the low-latency event stream generates a binary event image from short temporal intervals. Objects' region proposals (RPs) are created based on event image by Connected Component Labeling (CCL) [2], and the RPs are tracked based on neighboring events. Moderate-speed objects (<20 pixels/s) use grayscale image patches for NN classification, while high-speed objects (>20 pixels/s) leverage event-based motion trajectories, mitigating accuracy loss due to motion blur.

Fig. 2 presents the hardware architecture of our system, featuring an Event Signal Processor (ESP) for RP generation and Object Tracking (OT), an Image Signal Processor (ISP) for grayscale preprocessing, and a Neural Network Processing Unit (NPU) for classification. The ESP operates in an always-on power domain, activating the power-intensive NPU only when necessary. Within ESP, a frame builder reconstructs event images periodically, which are then denoised. Subsequently, Run Length Encoding (RLE) compresses adjacent events in each row into slices. The Region Proposal Unit (RPU) localizes and tracks objects, while the Fast Object Tracking Unit (FOTU) handles swift-moving objects.

The performance of the Object Detection and Tracking (ODT) greatly affects the reliability of the anti-UAV system. Event-based ODT incurs high dynamic power consumption as the system must remain continuously active to process every incoming event [3]. In contrast, ODT based on binary event frames reconstructed from event streams suffers from accuracy degradation when tracking rapidly moving objects [4-5]. Therefore, we proposed a hybrid ODT algorithm featuring both frame mode and event mode for efficient and fast object tracking (Fig. 3(top left)). Initially, the RPU operates in frame mode, merging adjacent slices from different rows into RPs. If an RP exceeds nine pixels in size, it is considered as a valid object, prompting the RPU to switch to event mode for tracking; otherwise, the RPU remains in frame mode to conserve power. In event mode, the RPU monitors incoming events around each RP, updating the RP's coordinates (RP*) based on the outermost event positions after receiving TH matched events. This event-driven approach enables tracking speeds up to 200 pixels/s. Additionally, the RPU periodically switches back to frame mode every 5 to 30 seconds to promptly detect new objects.

Through hardware-algorithm co-optimization, the RPU efficiently supports both frame-based and event-based tracking modes (Fig. 3, top right). When multiple PEs respond to the same input, their corresponding RPs are seamlessly merged by the PE controller. In both modes, each PE evaluates whether its buffered RP overlaps with the current input slice or event; upon a successful match, the PE's status register is set to one. During frame mode, the RP is updated immediately upon an overlap detection, whereas in event mode, updates occur only after the match counter surpasses an adaptive threshold (TH). This threshold scales with the object's size and velocity to maximize the intersection-over-union with ground truth. By sharing hardware between both modes, the hardware overhead of the RPU is reduced by 44.2% compared to the separate implementations. Besides, it significantly lowers processing latency compared to purely frame-based RP generation (Fig. 3(bottom left)), while achieving superior tracking accuracy against both CCL-based and event-driven SNN baselines [1] (Fig. 3(bottom right)).

Tracking high-speed UAVs introduces two main challenges—misaligned RPs and blurred RP regions in the grayscale image (Fig. 4, top left). To mitigate these issues, the Fast Object Tracking Unit (FOTU) is introduced (Fig. 4, bottom left). Specifically, thirty-two RP monitors concurrently observe variations in each PE's RP and adaptively adjust their RP update thresholds (TH). As illustrated in Fig. 4 (top right), whenever an RP update occurs in event mode, the corresponding PE's TH is recalibrated based on the object's size and velocity, enabling the RP to align closely with the ground-truth bounding box, thereby alleviating RP misalignment. To address motion blur, we propose classifying fast-moving objects based on their trajectories rather than blurred grayscale image patches. Experimental results on a drone dataset [2] demonstrate that trajectory-based classification achieves higher accuracy compared to patch-based methods when object speed exceeds 25 pixels/s. Consequently, trajectories of fast-moving objects are recorded point-by-point in the trajectory memory of the FOTU. To avoid redundant data storage and minimize memory usage, trajectory points are recorded only when the displacement step exceeds four pixels.

Fig. 5 shows the NPU for classifying tracked objects using either grayscale image clips or moving trajectories. The NPU employs a customized 64-bit instruction set to orchestrate end-to-end CNN inference, where the lowest three bits indicate operation types, and the remaining bits configure the various sub-blocks within the NPU. It integrates a 16×16 Processing Engine (PE) array, leveraging an output-stationary dataflow for efficient multiply-accumulate (MAC) operations, and each PE features zero-skipping capability to minimize power consumption. The proposed instructions enables all sub-blocks of the NPU to operate concurrently, significantly reducing processing latency compared to conventional CNN accelerators, which sequentially execute sub-functions. Additionally, the NPU supports data preloading to eliminate pipeline stalls between consecutive instructions. Given that the NPU accounts for over 50% of the chip's dynamic power, the proposed anti-UAV system activates the NPU only once per tracked object, thus preventing redundant classifications and substantially reducing power usage. Experimental evaluations demonstrate that this approach reduces the computational load of the NPU by 98.3% and 97% on drone and vehicle datasets, respectively.

The proposed anti-UAV system is fabricated using 40nm CMOS technology in a 2mm$^2$ chip area. It operates at 65~214MHz under a 0.6~1V supply. As shown in Fig. 6, it achieves the highest energy efficiency of 96pJ/(frame*pixel) and 61pJ/event at 0.8V and 153MHz under frame mode and event mode, respectively. The ODT accuracy is evaluated using public event camera-based drone [2] and vehicle [4] datasets. Compared with prior arts, our design realizes the most energy-efficient end-to-end ODT, attributed to its close coupled algorithm and hardware co-optimization. It achieves 98.2% average recognition accuracy for UAVs flying within 50~400m range and 5~80 pixels/s moving speed.

**Acknowledgement:**

This work was supported by the National Research Foundation, Singapore, under its Competitive Research Programme (NRF-CRP25-2020-0002), and in part by the Ministry of Education, Singapore, under its Academic Research Fund Tier 2 (MOE-T2EP50122-0024).

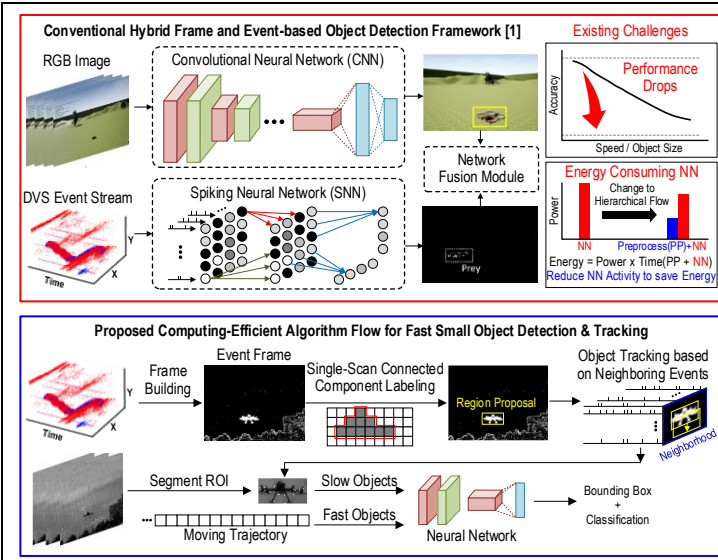

Fig. 1. Algorithm flow and challenges of existing anti-UAV systems (top) and the proposed algorithm flow (bottom).

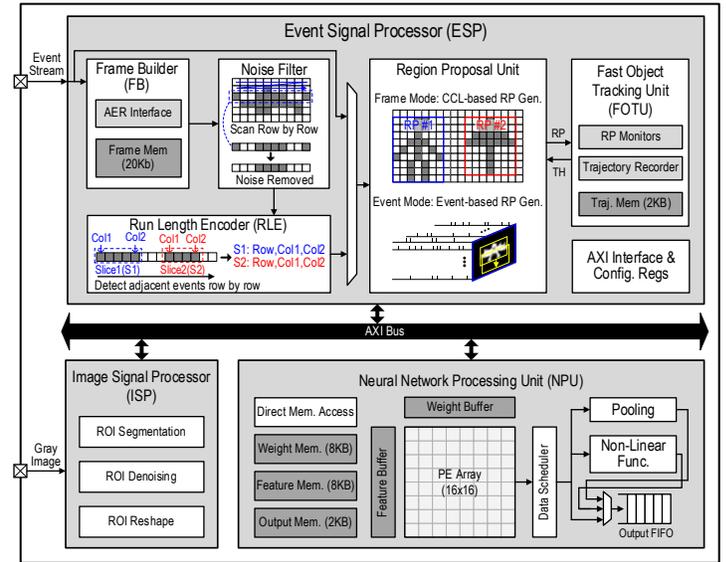

Fig. 2. Hardware architecture of proposed anti-UAV system.

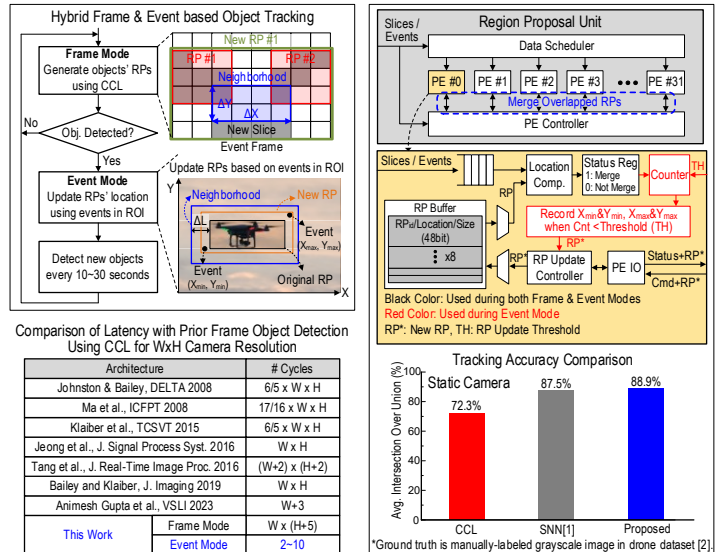

Fig. 3. Hybrid frame and event-based object tracking algorithm, architecture of the region proposal unit, and testing results.

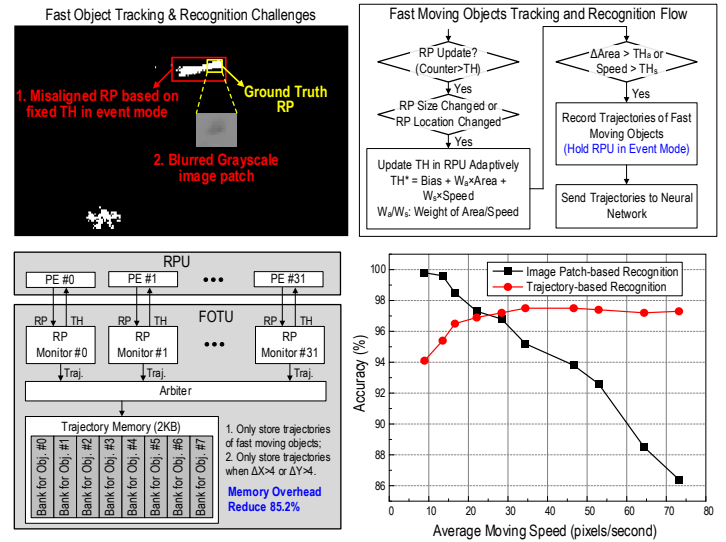

Fig. 4. Algorithm flow and architecture of fast object tracking unit, and object recognition accuracy under different moving speeds.

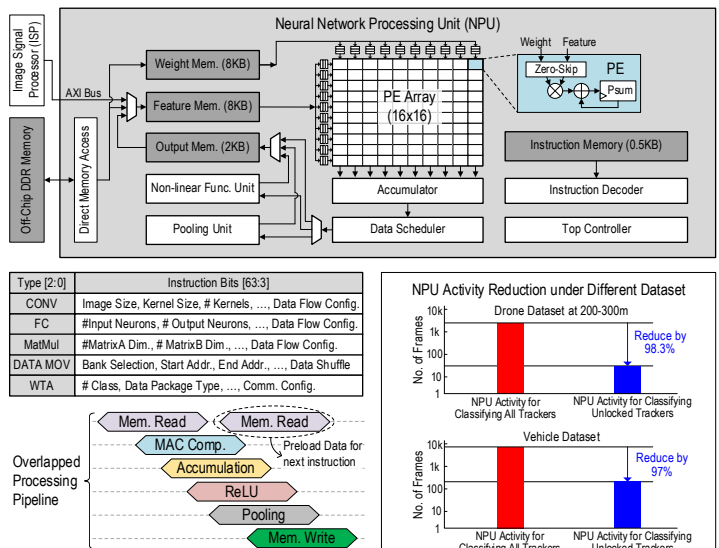

Fig. 5. Neural network processing unit with customized instruction set, overlapped processing pipeline, and activity gating scheme.

|  | ISSCC20 [6] | ISSCC23 [7] | ISSCC23 [1] | VLSI24 [5] | This Work |
|---|---|---|---|---|---|
| Technology | 65nm | 40nm | 40nm | 40nm | 40nm |
| Chip Area | 12mm² | 72.68mm² | 20.25mm² | 30mm² | 2mm² |
| Memory | 196KB | 2MB | 2.5MB | 1.7MB | 22.4KB |
| Frequency | 100MHz | 225MHz | 100MHz | 350MHz | 65-214MHz |
| Sensor Type | RGB | RGB | RGB + Event | Gray | Gray + Event |
| Resolution | 256×455 | 360×640 | 260×346 | 240×320 | 260×346 |
| Function[1] | RP | RP, OC | RP, OT, OC | RP, OC | RP, OT, OC |
| Peak Throughput | 44.2-67.1fps | 163.13fps | 11.1Meps | 30fps | Frame: 473.5fps Event: 10.25Meps |
| Power[2] | 7.3-99mW | 92.45mW | 4.6-21.3mW | 2.1mW | Frame: 519.6µW @60fps, 0.8V Event: 623.9µW @10.25Meps, 0.8V |
| Energy/Frame/Pixel | 212-1572.8pJ | 2.47nJ | NA | 1.16nJ | 0.096nJ |
| Energy/Event | NA | NA | 0.8nJ | NA | 0.061nJ |
| OT Accuracy | NA | NA | NA | NA | 82%@IoU=0.65[3a] 79%@IoU=0.65[3b] |
| OC Accuracy | NA | NA | 99%[3c] | NA | 99%[3a], 98.2%[3b] |

[1] Function: Region Proposal (RP), Object Tracking (OT), Object Classification (OC)
[2] Excluding external memory access power as other work.
[3] Accuracy tested on public vehicle dataset [4][3a], and drone dataset [2][3b].

Fig. 6. Comparison with prior arts.

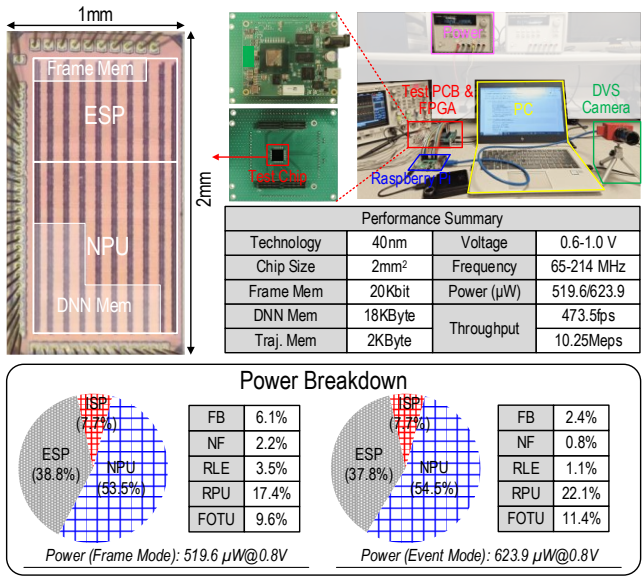

Fig. 7. Chip micrograph, testing setup, and testing results.